\documentclass[10pt,twocolumn,letterpaper]{article}

\usepackage{cvpr}              

\usepackage{bbm}
\usepackage{tikz}
\usepackage{comment}
\usepackage{graphicx}
\usepackage{amsmath}
\usepackage{amssymb}
\usepackage{booktabs}
\usepackage{algorithm}
\usepackage{algpseudocode}
\usepackage[accsupp]{axessibility} 
\usepackage{xfrac}
\usepackage{color}
\usepackage{wrapfig}
\usepackage{cite}
\usepackage{microtype}
\usepackage{tabularx}
\usepackage{booktabs}
\usepackage{multirow}
\usepackage{blindtext}
\usepackage{xspace}

\definecolor{dimgray}{rgb}{0.41, 0.41, 0.41}
\newcommand{\grc}[1]{\textcolor{dimgray}{#1}}

\usepackage[pagebackref,breaklinks,colorlinks]{hyperref}

\usepackage[capitalize]{cleveref}
\crefname{section}{Sec.}{Secs.}
\Crefname{section}{Section}{Sections}
\Crefname{table}{Table}{Tables}
\crefname{table}{Tab.}{Tabs.}

\def\confName{CVPR}
\def\confYear{2023}

\begin{document}

\title{Text2Scene: Text-driven Indoor Scene Stylization with Part-aware Details}

\author{Inwoo Hwang$^1$, Hyeonwoo Kim$^1$, and Young Min Kim$^{1, 2}$ 
\\
{\small $^1$Department of Electrical and Computer Engineering, Seoul National University}\\
{\small $^2$Interdisciplinary Program in Artificial Intelligence and INMC, Seoul National University}\\
}

\twocolumn[{%
\renewcommand\twocolumn[1][]{#1}%
\maketitle
\begin{center}
    \centering
    \captionsetup{type=figure}
    \includegraphics[width=\linewidth]{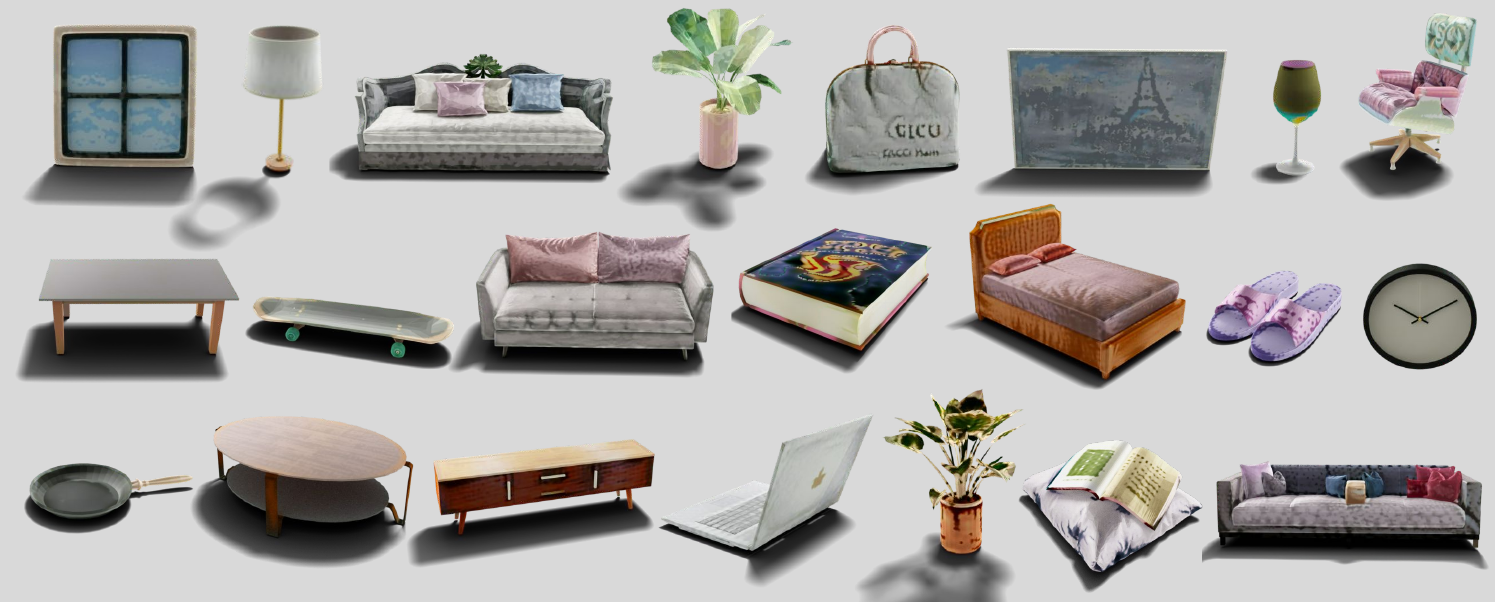}
    \captionof{figure}{Stylization results of diverse objects. Text2Scene creates more realistic and part-aware textures for various categories of 3D objects without any dedicated datasets for training, which can constitute high-quality virtual scenes.}
\label{fig:all object}
\end{center}%
}]

\begin{abstract}

We propose Text2Scene, a method to automatically create realistic textures for virtual scenes composed of multiple objects.
Guided by a reference image and text descriptions, our pipeline adds detailed texture on labeled 3D geometries in the room such that the generated colors respect the hierarchical structure or semantic parts that are often composed of similar materials.
Instead of applying flat stylization on the entire scene at a single step, we obtain weak semantic cues from geometric segmentation, which are further clarified by assigning initial colors to segmented parts.
Then we add texture details for individual objects such that their projections on image space exhibit feature embedding aligned with the embedding of the input.
The decomposition makes the entire pipeline tractable to a moderate amount of computation resources and memory.
As our framework utilizes the existing resources of image and text embedding, it does not require dedicated datasets with high-quality textures designed by skillful artists.
To the best of our knowledge, it is the first practical and scalable approach that can create detailed and realistic textures of the desired style that maintain structural context for scenes with multiple objects.

\end{abstract}
\begin{figure*}[t!]
\centering
\includegraphics[width=\linewidth]{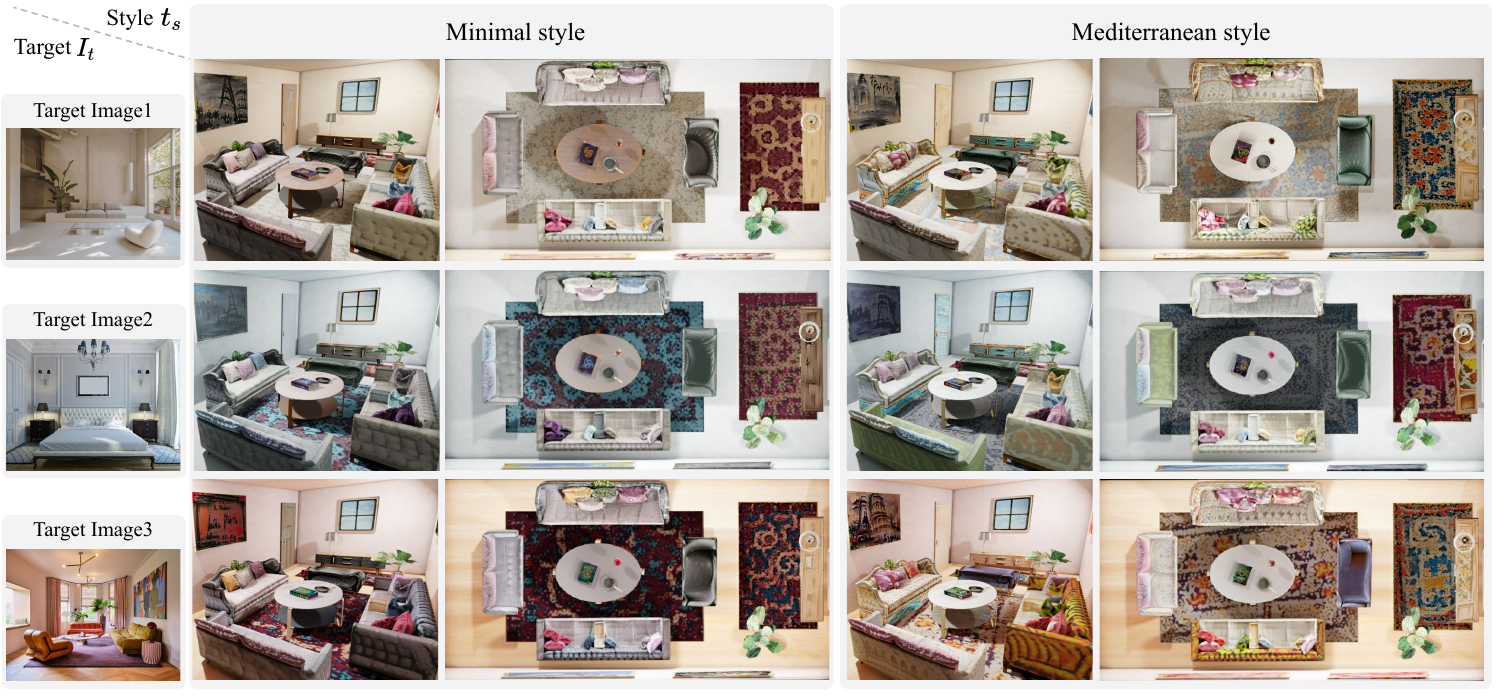}
\vspace{-2.0em}
\caption{Our scene stylization results. Given a target image $I_t$ and the style text $t_s$, Text2Scene can produce the stylized results for the entire scene.}
\label{fig:scene results}
\vspace{-0.5em}
\end{figure*}

\section{Introduction}
\label{sec:intro}

Virtual spaces provide an immersive experience for metaverse, films, or games.
With increasing demands for virtual environments, various applications seek practical methods to create realistic 3D scenes with high-quality textures.
Currently, skillful artists need to manually create 3D assets and accompanying textures with careful parameterization, which is not scalable enough to account for the diverse content the industry is heading for.
Scenes can also be populated with existing 3D database models or created with recent shape-generation approaches using data-driven methods~\cite{Paschalidou2021atiss,zhang2022cgca}.
However, most of them lack texture information or are limited to simple coloring.

To build realistic content, we need fine details containing the artistic nuances of styles that obey the implicit correlations with geometric shapes and semantic structure.
Recent works provide methods to color a single object with the help of differentiable rendering~\cite{chen2019_dibr}, but often they are limited to single texture or blurred boundaries~\cite{Michel2021text2mesh,chen2022tango, yin2021_3DStyleNet, mishra2022clipnnst, khalid2022clipmesh}.
More importantly, the 3D objects are often textured in isolation, and only limited attempts exist to add visual appearances for large-scale scenes with multiple objects~\cite{hao2021GANcraft,hollein2022stylemesh,Jeong2022indoorstyle}.
The biggest challenge is adding consistent style for an entire scene, but still accounting for the boundaries of different materials due to the functional and semantic relationship between parts, as observed within real-world scenes.

Our proposed Text2Scene adds plausible texture details on 3D scenes without explicit part labels or large-scale data with complex texturing.
We take inspiration from abundant 3D shape and image datasets and decompose the problem into sub-parts such that the entire scene can be processed with a commodity memory and computation.
Given scenes of multiple objects of 3D mesh geometry, we separately handle walls and individual objects.
Specifically, the stylization of walls is formulated as texture retrieval, and the objects are initialized with base colors.
From the base color assignment, we can deduce the part-level relationship for stylization and further refine them in later stages, such that their rendered images are close to the input text within the joint embedding space of foundational models.

Our coarse-to-fine strategy keeps the problem tractable yet generates high-quality texture with clean part boundaries.
We first create segments of input mesh such that the segment boundaries align with low-level geometric cues.
Then we start with the simplified problem of assigning a color per segment.
Interestingly, the prior obtained from large-scale image datasets assign similar colors for the parts with similar textures, reflecting the semantic context or symmetry as shown in Figure~\ref{fig:all object}.
We add the detailed texture on individual objects as an additional perturbation on the assigned base colors by enforcing constraints on the image features of their projections.
The additional perturbations are high-frequency neural fields added to the base color.

In summary, Text2Scene is a new method that 
\begin{itemize}
    \item can easily generate realistic texture colors of the scene with the desired style provided by text or an image;
    \item can add detailed texture that respects the semantic part boundaries of individual objects; and
    \item can process the entire scene without a large amount of textured 3D scenes or an extensive memory footprint.
\end{itemize}
We expect the proposed approach to enable everyday users to quickly populate virtual scenes of their choices, and enjoy the possibility of next-generation technology with high-quality visual renderings.

\begin{figure*}[t!]
\centering
\includegraphics[width=\linewidth]{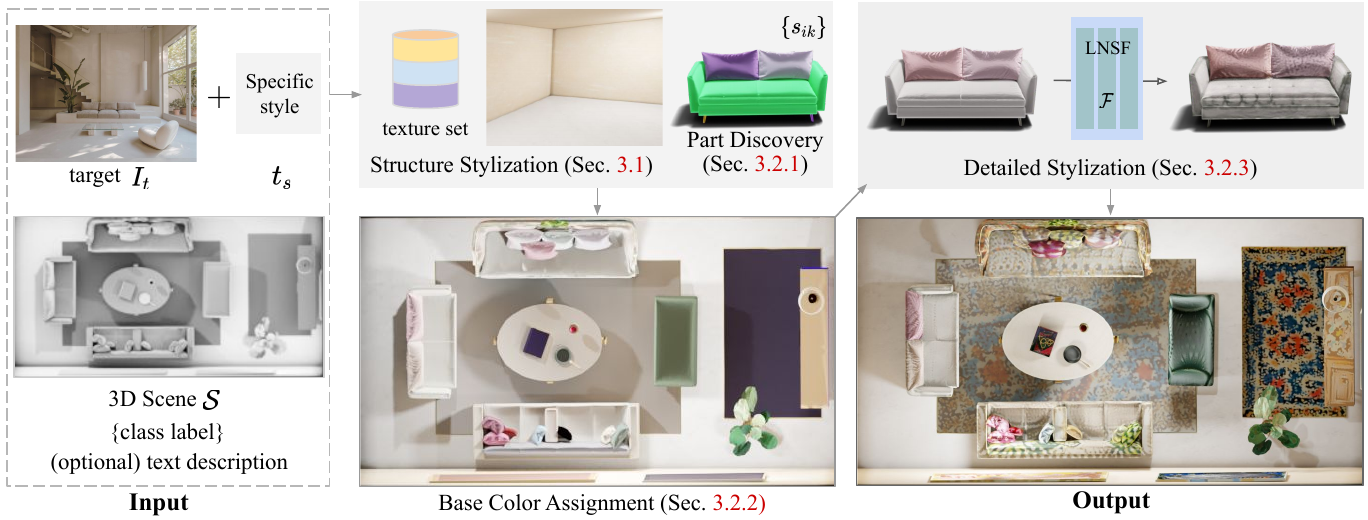}
\caption{Overall pipeline of Text2Scene. 
Given a 3D scene $\mathcal{S}$ with optional text description, we generate textures guided by a target image $I_t$ and the appearance style provided as a text $t_s$.
We first stylize the structure by texture retrieval, and each object is pre-processed for part discovery for stylization.
Then, we assign base colors to object parts and add local details for each object with the designated LNSF.}
\label{fig:pipeline}
\vspace{-0.9em}
\end{figure*}

\section{Related works}
\label{sec:related}

\paragraph{3D Shape Understanding and Segmentation}
Our proposed method creates a realistic texture that abides by implicit rules of the material composition of different object parts.
Different materials or textures are often assigned to different objects' functional parts, whose boundaries align with geometric feature lines.
A handful of previous works find segments that constitute a 3D model with geometric information~\cite{Katz2003MeshDecompose,chen2020bspnet,deng2020cvxnet,Paschalidou2021NERUALPARTS}, while others train 3D features to distinguish part labels provided in datasets~\cite{Mo2019partnet,qi2017pointnetplusplus,chen2019baenet}.
Recently, PartGlot~\cite{koo2022partglot} suggested discovering part segments from language references, which should contain functional information.
However, the geometric or functional distinction does not always clarify the texture boundaries, with possible additional diversity originating from the designers' choice.
The intricate rules are multi-modal distributions composed of a mixture of discrete part assignments and continuous texture details, whose results are contained within visual datasets of scenes. 

\paragraph{Neural Fields and Texture}
Traditional texture atlas of 3D geometry is represented as a mapping from a planar image to a manifold embedded in 3D space and involves complex parameterization.
With the increasing popularity of neural representation, TextureFields~\cite{Oechsle2019texturefield} represented texture as a mapping from 3D surface points to RGB color values without separate parameterization.
It is deeply connected with graphics pipelines that adapt coordinate-based functions to depict SDF shapes ~\cite{Mescheder_2019_occnet,https://doi.org/10.48550/arxiv.1901.05103,https://doi.org/10.48550/arxiv.2006.09661} or novel-view synthesis using implicit volumes\cite{mildenhall2020nerf}.
Various works show neural implicit representation is highly flexible and free from domain structure or resolution~\cite{wang2021clipnerf,kobayashi2022distilledfeaturefields,martinbrualla2020nerfw,pumarola2020dnerf,Azinovic_2022_CVPRrgbd}.
Recently, Text2Mesh~\cite{Michel2021text2mesh} generated the deformation and coloring of input mesh with neural fields guided by the joint embedding of rendered image and input text.
Interestingly, the generated mesh deformation and coloring also contain semantic part information.
Our Text2Scene framework further extends the ability and observes distinct part boundaries, which could not be captured in previous works.
We also increase the reality of resulting scenes with high-frequency details.
We encode the input to the neural network with a high-order basis of intrinsic features~\cite{Koestler2022intrinsicneuralfields} and apply the coarse-to-fine strategy as shown with geometric details of Yifan \textit{et al.}~\cite{yifan2022geometryconsistent}.

\paragraph{Text-Driven 3D Stylization}
Recently neural networks trained with large-scale images and texts demonstrated powerful performance in many tasks with their extensive representation power.
Here we primarily focus on 3D stylization attempts using image features.
CLIP~\cite{2021CLIP} learns latent space with a large amount of image text pairs, and the additional text input allows semantic manipulation for various generative tasks, including images~\cite{2022dalle2, rombach2021stablediffusion, Patashnik2021styleclip, ruiz2022dreambooth, chen2022text2light}, videos~\cite{bar2022text2live, hong2022cogvideo}, motions~\cite{tevet2022motionclip, tevet2022humanmotiondiffusion}, and 3D assets ~\cite{jain2021dreamfields, poole2022dreamfusion, khalid2022clipmesh,sanghi2021clipforge}.
However, for text-driven 3D stylization, it is still difficult to define clean texture boundaries and correlation with instance-level subtleties~\cite{Michel2021text2mesh, chen2022tango, mishra2022clipnnst} or focus only on the specific type, such as human~\cite{youwang2022clipactor, hong2022avatarclip}.
As another way to stylize a 3D scene~\cite{magicdecorator,Yeh2022photoscene}, Yeh \textit{et al.}~\cite{Yeh2022photoscene} matches the input image features with CAD input using differentiable material graphs~\cite{shi2020match}.
However, the representation is inherently limited to a combination of material libraries and cannot handle non-homogeneous details such as painting.
On the other hand, our approach can generate texture beyond the repetitive low-level patterns of input materials and introduce part-aware texture with fine details that maintain geometric and semantic consistency.

\section{Method}

We stylize a 3D indoor scene without sophisticated techniques or software tools for 3D modeling or texturing.
The input 3D scene $\mathcal{S}=\{\mathcal{W}, \mathcal{O}\}$ is  a set of structure components $\mathcal{W}$ and a set of objects $\mathcal{O}$.
The structural components $\mathcal{W}$ are walls, ceilings, and floors, whereas the objects $\mathcal{O}=\{M_{i}\}$ are the 3D mesh models $M_i$.
We assume that all the components have their corresponding class labels and optionally have text descriptions.

The desired color distribution is provided as a target image $I_t$, which could be retrieved from the web. 
Also, a specific appearance style description $t_{s}$ is provided as input to enforce style consistency among a set of objects $\mathcal{O}$. 
The color distribution compares the color histogram of the input target image $I_t$ against the rendering of the current stylized scene $I$.
Specifically, the histogram loss $\mathcal{L}_\text{hist}$ is defined as below
\begin{equation}
\mathcal{L}_\text{hist}\left( I,I_{t}\right) = \| H^{1/2}\left( I\right) -H^{1/2}( I_{t}) \| _{2},
\label{eq:hist}
\end{equation}
where $H\left( \cdot \right) $ indicates differentiable color histogram operator~\cite{afifi2021histogan}.
Also, we augment the losses derived from the joint embedding of text and images  \cite{2021CLIP} by generating text descriptions $T$ of the context using semantic labels and comparing them against the rendered image $I$.
If we denote the pre-trained encoder for image and text as $E_1$ and $E_2$, respectively, the CLIP similarity loss is defined as
\begin{equation}
\mathcal{L}_\text{clip}\left( I,T\right) = 1 - sim(E_{1}\left(I\right),E_{2}\left(T\right)),
\label{eq:clip}
\end{equation}
where $sim\left( \mathbf{x},\mathbf{y}\right) =\dfrac{\mathbf{x}^\top\mathbf{y}}{\left\| \mathbf{x}\right\| _{2}\left\| \mathbf{y}\right\| _{2}}$ is the cosine similarity.

The overall pipeline is described in Fig.~\ref{fig:pipeline}. 
We obtain the texture for the structure $\mathcal{W}$ by texture retrieval, which is described in Sec.~\ref{sec:structure}.
The objects are stylized with additional decomposition to respect local part boundaries and, simultaneously, to practically handle multiple entities with details (Sec.~\ref{sec:object}).

\subsection{Structure Stylization}
\label{sec:structure}

We assign one coherent texture per structural element of $\mathcal{W}$.
Compared to objects, the structural elements, such as walls or ceilings, are of simple planar geometry, and their textures are not heavily dependent on the relationships between different functional parts.
For structural elements, it suffices to pick texture from the texture set of an existing material library of MATch~\cite{shi2020match}, which contains homogeneous material.
If we instead utilize visual features or CLIP embeddings for them, the resulting stylization exhibits undesired artifacts of various sizes instead of constant patterns, as shown in the supplementary.

We randomly initialize the texture from the texture set and render the structure image $I_s$, a bare room only containing the structural elements $\mathcal{W}$ without objects.
Then the materials are compared to the target image $I_t$ for the histogram losses of Equation (\ref{eq:hist}).
The additional text prompt $T_s$ is given as `a structure of a room' to provide the context with $\mathcal{L}_\text{clip}\left( I_{s},T_{s}\right)$.
In summary, the texture of the structural element is retrieved to have the lowest score on the following criteria:
\begin{equation}
\mathcal{L}_\text{hist}\left( I_{s},I_{t}\right)+
\lambda _{1}\cdot\mathcal{L}_\text{clip}\left( I_{s},T_{s}\right).
\label{eq:structure}
\end{equation}

\subsection{Object Stylization}
\label{sec:object}

Object stylization involves understanding the semantic structure hidden behind the mesh representation.
As a pre-processing, we first subdivide individual objects $M_i$ in $\mathcal{O}$ into part segments $\{s_{ik}\}$ as described in Sec.~\ref{sec:object_part}.
Then the scene is stylized in two steps.
First, we assign base colors into individual parts to minimize the style loss for the entire scene $\mathcal{S}=\{\mathcal{W}, \mathcal{O}\}$ (Sec.~\ref{sec:object_base}).
Here we are optimizing for the discrete set of colors assigned to subdivided parts obtained from the pre-processing.
Then the textures for individual objects are further optimized to generate fine details (Sec.~\ref{sec:object_detail}).

\subsubsection{Part Discovery for Object Stylization}
\label{sec:object_part}

We first decompose individual objects into parts such that each part is composed of the same material or texture.
The distinctive part boundaries are critical in providing semantic consistency and, therefore, perceptual realism toward the scene.
Similar to 3D part segmentation methods, we first find super-segments based on geometric features which provide the granularity to define textural parts.
For a given 3D object mesh $M_{i}$, the initial segments $\{s_{ik}^{0}\}$ are the decomposition applying the method by Katz \textit{et al.}~\cite{Katz2003MeshDecompose}.
The decomposition is designed to be an over-segmentation of our aim.
We generate a graph $\mathcal{G}_{i}^{0}$ where each node is the segment and edges connect neighboring segments.
Then we incrementally merge segments that belong to the same texture until convergence.
Note that, our part discovery method operates robustly regardless of the initial composition, but we use ~\cite{Katz2003MeshDecompose} which preserves the original geometry and details.

\begin{figure}[t!]
\centering
\includegraphics[width=\linewidth]{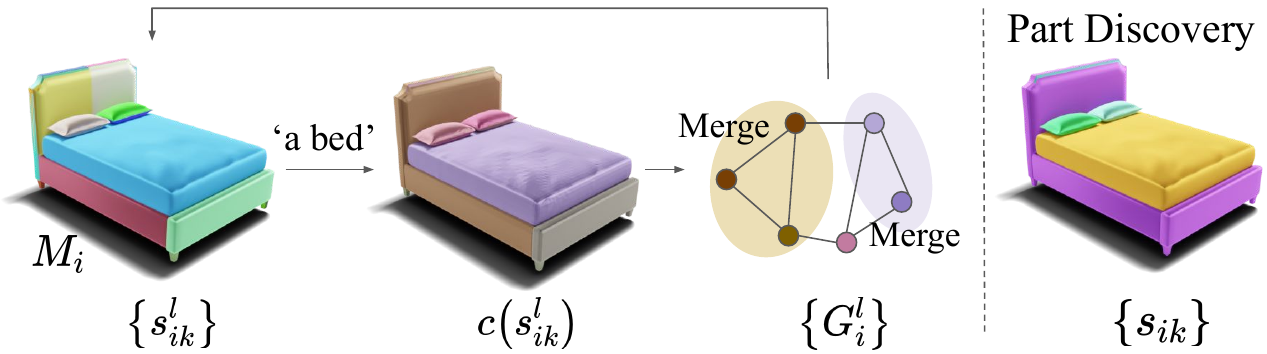}
\caption{Overall pipeline of part discovery. 
We discover parts for object stylization from the super-segments of 3D object mesh.
Given a 3D object mesh $M_{i}$ with segments $\{s_{ik}^{l}\}$ at $l^\text{th}$ \text{iteration}, we assign a color $c(s_{ik}^{l})$ per segment and generate a graph $\mathcal{G}_{i}^{l}$.
The pair of neighboring nodes is merged if the distance between assigned colors $c(s_{ik}^{l})$ is within a threshold, then, move to $(l+1)^\text{th}$ \text{iteration}.
}
\vspace{-0.9em}
\label{fig:part_discovery}
\end{figure}

The challenge here is that, unlike semantic segmentation approaches, no large-scale public dataset exists that provides the ground truth for `texture similarity' as the segmentation labels.
We find a supervision signal from the large-scale pre-trained model, and create a simple text prompt $T_{i,c}$ using the class name, such as \textit{a bed} or \textit{a chair}.
At $l^\text{th}$ \text{iteration}, we assign a color $c(s_{ik}^{l})$ to each segment $\{s_{ik}^{l}\}$, which is optimized to minimize the distance between the rendered images $I_{M_i}$ of multiple viewpoints and the text $T_{i,c}$ in the joint embedding space of CLIP, or $\mathcal{L}_\text{clip}(I_{M_i}, T_{i,c})$ as defined in Equation (\ref{eq:clip}).
If the resulting colors assigned to two adjacent segments are similar, the two parts are likely to be the parts with the same texture source.
Therefore we merge the two segments for the next iteration $\{s_{ik}^{l+1}\}$.
In particular, we merge segments if the assigned color has a distance of less than a threshold $\lambda_{th}$ in the CIE color space which is known to
be related to human perception.
Note that, while the initial color is gray for the assignment, merging segments happen with the optimized color.
We repeat the process until the number of segments does not decrease anymore and empirically found that it usually converges within 2-3 steps.
The overall pipeline for part discovery is also described in Fig.~\ref{fig:part_discovery}.

\subsubsection{Part-level Base Color Assignment}
\label{sec:object_base}

After the objects  $M_{i} \in \mathcal{O}$ are decomposed into parts $\{s_{ik}\}$, we assign a solid color per part, namely $c\left( s_{ik}\right)$.
The base color assignment handles a low-dimensional optimization space with a coarse set of discrete parts but still observes the holistic distribution of the entire scene.
Then the base color is combined with the output of designed neural style fields in Sec.~\ref{sec:object_detail}, which generate high-frequency local texture.

We optimize the base color using the combined loss as before, $\mathcal{L}_\text{color,scene} + \mathcal{L}_\text{clip,scene}$.
The color loss again considers the similarity with the target image $\mathcal{L}_\text{hist}\left( I,I_{t}\right)$.
The scene being optimized $I$ is rendered with the stylized structure $\mathcal{W}$ and the current estimates of the base colors.
The clip loss considers both individual objects and the global scene and is calculated as the sum of object clip loss and global clip loss.
\begin{equation}
\mathcal{L}_\text{clip,scene} = 
\lambda _{2}\cdot\sum _{i}\mathcal{L}_\text{clip}\left( I_{M_{i}},T_{i}\right) +
\lambda _{3}\cdot \mathcal{L}_\text{clip}\left( I,T\right).
\label{eq:style_Scene}
\end{equation}
Unlike Sec.~\ref{sec:object_part}, text description $T_i$ for object $M_{i}$ could be a simple text prompt using the class name, or a detailed text prompt based on user choice.
We render individual objects from various angles $I_{M_i}$ and compare them with the text description $T_i$.
The embedding for the scene is also compared against the text embedding $T$ represents the type of scene, for example, `a bedroom'.
By jointly applying the loss, the base color $c\left( s_{ik}\right)$ is selected as a representative color that harmonizes nicely with the global context.

\subsubsection{Detailed Stylization}
\label{sec:object_detail}

The base color is combined with additional details regressed from a neural network to express the detailed local texture.
We define \emph{local neural style field (LNSF)} for each object, which generates the local textures added to the base color.
The color for point $p$ is defined by the following equation,
\begin{equation}
c\left( s_{p}\right) + \alpha \cdot \mathcal{F}_i\left( \gamma\left( p\right), \phi \left( p\right), s_{p}\right).
\label{eq:detail}
\end{equation}
We train a LNSF $\mathcal{F}_i$ per object, which outputs the color to be added to the base color $c(s_p)$, where $s_p$ indicates the part segment id.
The color range $\alpha$ maintains the final color to be similar to the base color.
$\gamma\left(\cdot \right)$ is the positional encoding of the $xyz$ coordinate to capture high-frequency details, and  $\phi (\cdot)$ represents the coefficients of the eigenfunctions for the intrinsic geometry using the Laplace-Beltrami operator.
Therefore the object-specific neural fields respect the part boundaries and local geometric details.

\begin{figure*}[t!]
\centering
\includegraphics[width=0.98\linewidth]{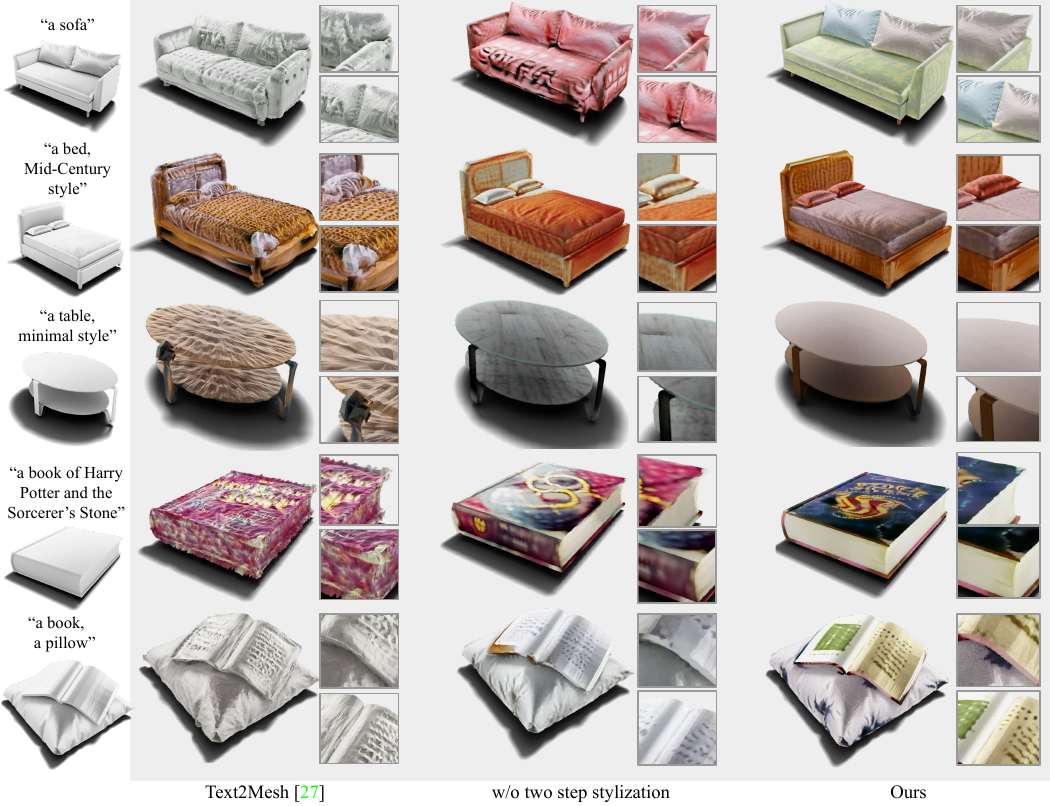}
\caption{Visual comparison with baselines. Our method utilizes the found part information and generates detailed textures through the designed network in combination with a coarse-to-fine stylization scheme.
As a result, it generates a globally harmonious 3D style that clearly classifies part segments.
We can handle diverse text such as simple categories or styles, detailed descriptions, or text that includes different two object categories.}
\label{fig:object}
\vspace{-0.63em}
\end{figure*}

LNSF for each object are trained with additional style context $\mathcal{L}_\text{clip}( I_{M_{i}},T^{+}_{i})$.
The text prompt $T^{+}_{i}$ augments the object description of $T_i$ with appearance style description $t_{s}$, such as `minimal style' or `Mid-Century style'.
By enforcing the same appearance style description, we can weakly bind the styles of individual objects in the same scene while optimizing separately.
We could also use $T^{+}_{i}$ instead of $T_{i}$ for optimizing the base color, and it shows slightly better results.

\paragraph{Part-aware Geometric Deformation}

Even though our main focus is to add colors to objects with local part-aware details, we could also concurrently produce part-aware geometric deformation with the same architecture.
We can slightly change the LNSF $\mathcal{F}$ to have two branches of output that estimate color and displacement.
For each point on the surface, we assign color by Eq.~\ref{eq:detail} and adjust displacement along the vertex normal direction.
To learn the effective deformation, we add geometric loss $\mathcal{L}_\text{clip}( I^{geo}_{M_{i}},T^{+}_{i})$, where $I^{geo}_{M_{i}}$ is an image rendering textureless geometry as~\cite{Michel2021text2mesh}.

\paragraph{Rendering and Implementation Details}
To render each object, the individual objects $M_{i}$ are scaled to fit a unit box.
The predicted color of each vertex allows the entire mesh to be differentiable rendering through interpolation using~\cite{chen2019_dibr}.
We render individual objects with random augmented backgrounds (white, black, random Gaussian, chess board), which helps the pipeline focus on the foreground object~\cite{jain2021dreamfields,hong2022avatarclip}.
Inspired by~\cite{poole2022dreamfusion}, since CLIP has a bias for the canonical pose, we augment the text prompt with the azimuth angle of view, namely `front view of', `side view of' or `back view of'.
Finally, in Sec~\ref{sec:object_detail}, random perspective transformation and random crop boost learning local details.
To render a scene, we randomly sampled from pre-define 20 camera poses to evenly cover the entire scene $\mathcal{S}$.

\section{Experiments}

Our stylization is first evaluated for individual objects in Sec.~\ref{sec:object results}. We evaluate the quality of textured object meshes and also assess that our pipeline can discover part segments to assign realistic stylization.
Then we demonstrate the stylization results of room-scale scenes with multiple objects in Sec.~\ref{sec:scene_results}.

\subsection{Object Stylization}
\label{sec:object results}

We show that our method can stylize various objects and produce realistic 3D assets to populate virtual scenes.
We use object meshes of various types and sizes, collected from Turbo Squid~\cite{TurboSquid}, 3D-FUTURE~\cite{fu20213dfuture}, and Amazon Berkeley Objects~\cite{collins2022aboamazon}.
For detailed stylization, we subdivide the object into an average of 119769 faces and 60437 vertices. 
For large general objects such as beds, sofas, tables, etc., we use class labels and the text input for the specific style as `a [\textit{class label}], [\textit{specific}] style'.
However, we need a detailed explanation for small objects that cannot be explained only by class labels, such as books.
For these objects, we configure detailed text individually, for example, `a book of Harry Potter and the Sorcerer's Stone'.

Figure~\ref{fig:object} shows the renderings of stylized results.
We render objects in Blender~\cite{Blender} with a fixed lighting setup.
The input mesh and the text are also provided in the left column.
Since our method utilizes the obtained part information and coarse-to-fine stylization scheme with a two-step approach, it creates a more realistic texture with clear boundaries for each part of the 3D assets.
Text2Mesh~\cite{Michel2021text2mesh} is another text-driven 3D stylization approach, which adds the RGB color and deformation fields on the vertices of mesh.
While it augments the detailed variations on the input mesh, the deformation map can occasionally introduce undesired artifacts and the part boundaries are only roughly estimated.
We also provide the results of an ablated version that directly uses LNSF without a two-step stylization scheme.
Our part discovery module and coarse-to-fine stylization scheme play a critical role in producing realistic assets with high-quality stylization.
Figure~\ref{fig:all object} contains more results on diverse objects.

\begin{figure}[ht!]
\centering
\includegraphics[width=0.9\linewidth]{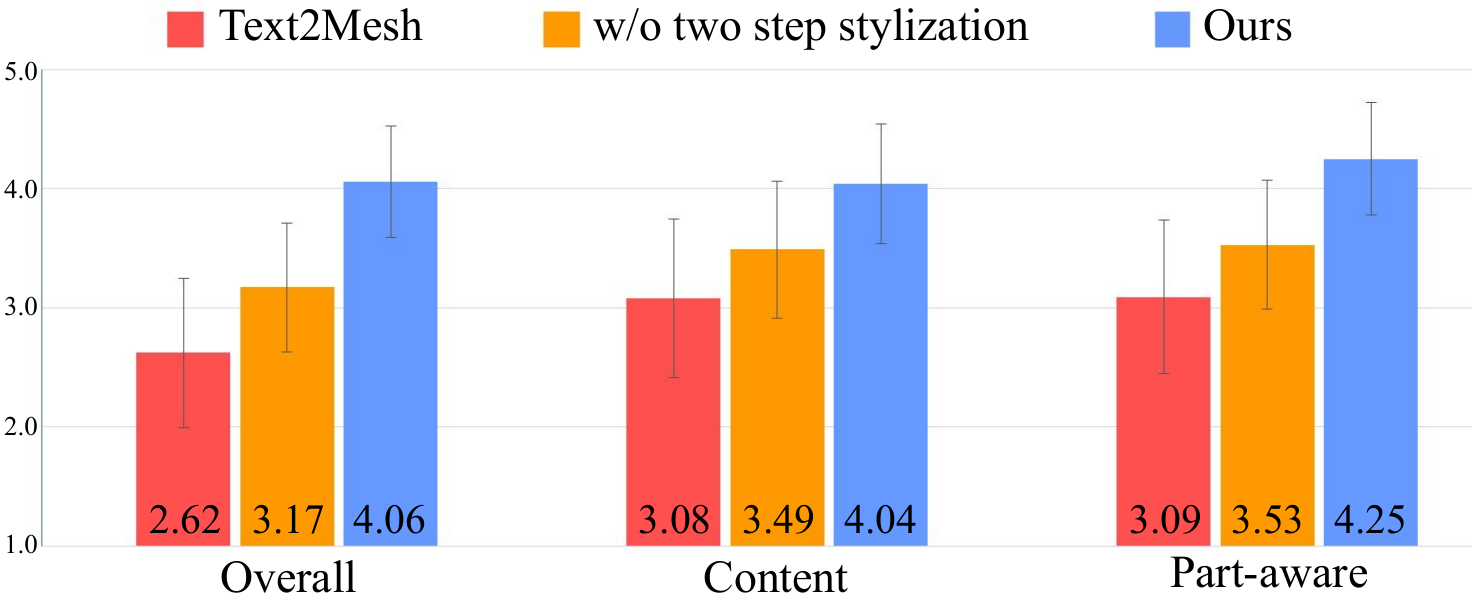}
\caption{Results of user study for object stylization}
\label{fig:user study}
\vspace{-1.5em}
\end{figure}

Because stylization is a subjective task, and no ground truth or metrics exist, we conduct a user study for quantitative evaluation.
We ask 98 users to rate the quality of generated outputs on a scale of 1 (worst) to 5 (best) in response to the following questions:
(Q1) ‘How natural are the output results?’, (Q2) ‘How well does the output contain text information?’, and (Q3) ‘How well does the output reflect part information?'
Figure~\ref{fig:user study} contains the mean and the standard deviation of the scores.
Our method outperforms competing methods in all aspects.
Therefore we conclude that our method generates a realistic texture that abides by input text description and semantic part information.

\begin{figure}[h!]
\centering
\includegraphics[width=0.95\linewidth]{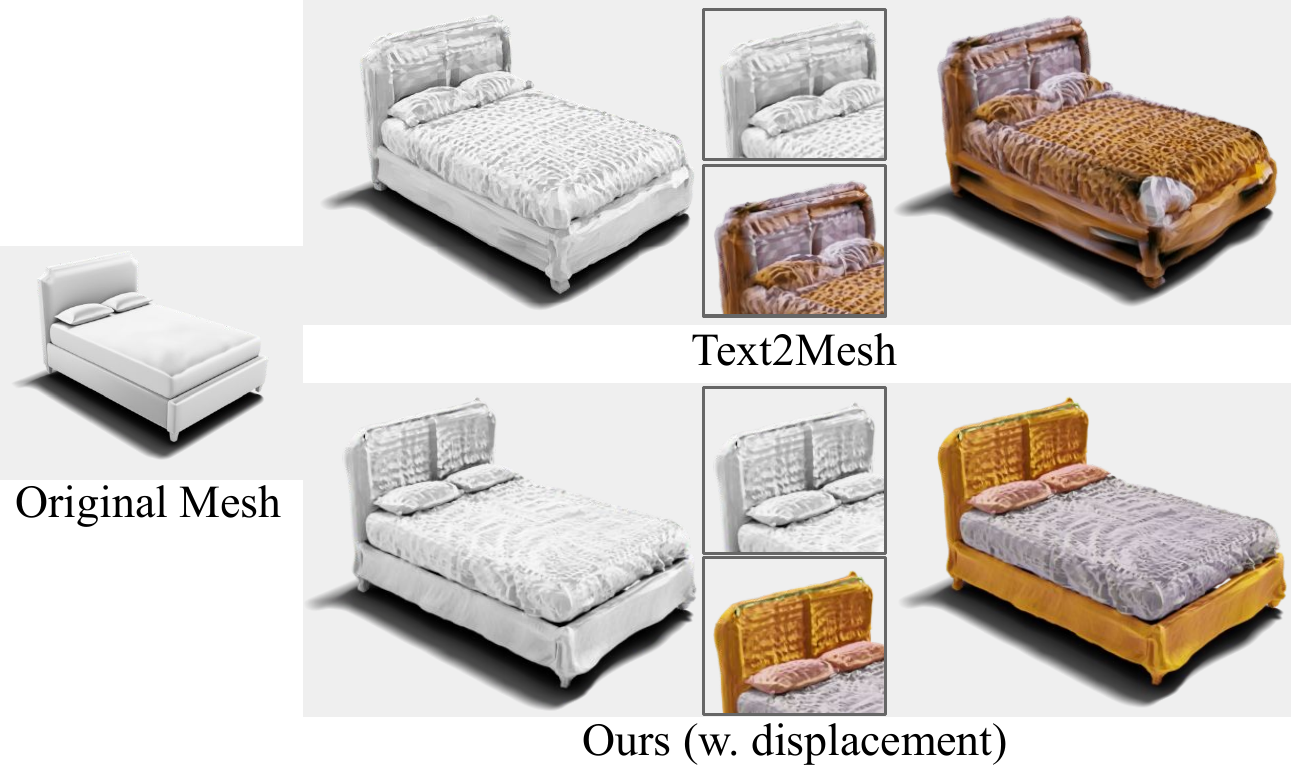}
\caption{With a slightly modified LNSF $\mathcal{F}$, we can simultaneously generate the texture with both color and displacement fields, and all of the generated style information respects the part information discovered.}
\label{fig:with displ}
\vspace{-1.2em}
\end{figure}

We also provide a parallel comparison against Text2Mesh by incorporating deformation fields with our approach.
Our original method preserves the original shape of the mesh, while Text2Mesh deforms vertices along the normal direction in addition to generating texture.
We can make a similar version of our LNSF and produce additional deformation as described in Sec.~\ref{sec:object_detail}.
Figure~\ref{fig:with displ} shows the deformed results from the source mesh compared to Text2Mesh.
Since our approach explicitly considers the part information with the two-stage approach, our results with deformation fields also respect different semantic parts of the object.

\begin{figure}[h!]
\centering
\includegraphics[width=1.0\linewidth]{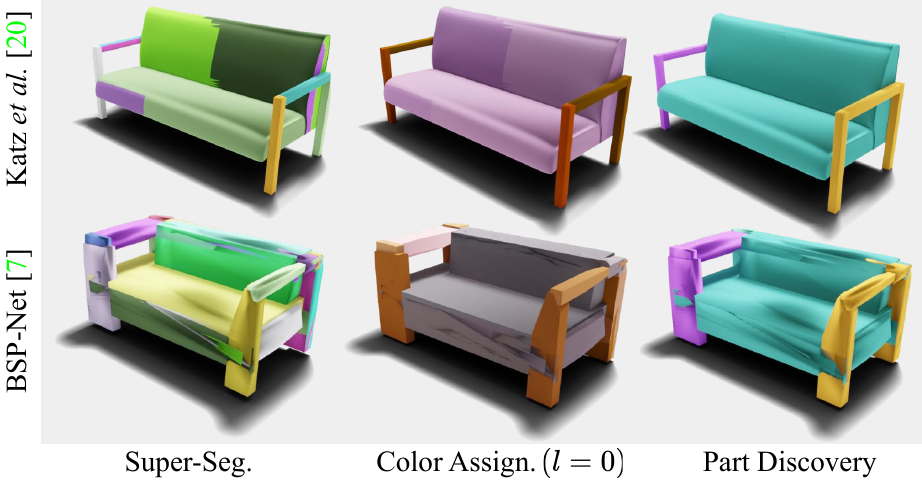}
\caption{The robustness against initial super-segmentation. The super-segments of the top and the bottom rows are generated by~\cite{Katz2003MeshDecompose} and~\cite{chen2020bspnet}, respectively (left). 
Starting from the initial color assignment (middle), our approach stably finds part decomposition to assign different base colors and therefore different texture information (right). The text \textit{a chair} is used.
}
\vspace{-0.9em}
\label{fig:bsp part discovery}
\end{figure}

\paragraph{Part Discovery}
\label{sec:part discovery}
As a side product of object stylization, we can discover different parts with distinguished boundaries that can guide a realistic color assignment (Sec.~\ref{sec:object_part}).
Our part discovery combines geometric super-segments and implicit clues from image-text embeddings.
The initial super-segments guide the algorithm to follow geometric feature lines, and the final results stably find part information despite different initializations.
Figure~\ref{fig:bsp part discovery} shows examples initialized with Katz \textit{et al.}~\cite{Katz2003MeshDecompose} (top) and BSP-Net~\cite{chen2020bspnet} (bottom).
Even for input mesh with bad topology or different initialization, our iterative part discovery quickly finds visually coherent parts without any training with segmentation labels.

\subsection{Scene Stylization}
\label{sec:scene_results}

Now we demonstrate that our text2scene framework can quickly generate a realistic texture for a room with multiple objects.
Recall that the 3D scene $\mathcal{S}=\{\mathcal{W}, \mathcal{O}\}$ is composed of the structure components $\mathcal{W}$ and a set of objects $\mathcal{O}$.
We use the same objects as described in Sec.~\ref{sec:object results} and arrange them to constitute scenes.
As there is no existing dataset composite of complete object meshes with labels, we built a total of four scenes: two bedrooms and two living rooms.
Each scene contains an average of 20 objects of various sizes and classes.
Additionally, we provide a target image $I_t$ for the color distribution, and a text prompt describing the desired style $t_s$.
While it can be daunting to define the desired style for the entire scene, images and texts can provide a simple way to deliver the information.

Figure~\ref{fig:scene results} shows stylized results of the same geometry, but observing various target images and style prompts.
The generated textures respect the semantic labels of various furniture and different parts and contain localized diverse details.
This is in contrast to many prior stylization methods where the fine perturbations are spread throughout the scenes.
We also show scenes with different types of rooms containing different objects, but stylized based on the same target image in Fig.~\ref{fig:various room}.
Additional results of various input configurations are available in the supplementary material.
Note that our target image $I_t$ does not need to restrict as an indoor image, and can be replaced such as natural photographs.
And by changing random seeds, diverse results can be obtained from the same input conditions.
Also, since we stylize the entire object, we can easily edit the 3D scene through object relocation.
These results can also be found in the supplementary.

\begin{figure}[h!]
\centering
\includegraphics[width=0.9\linewidth]{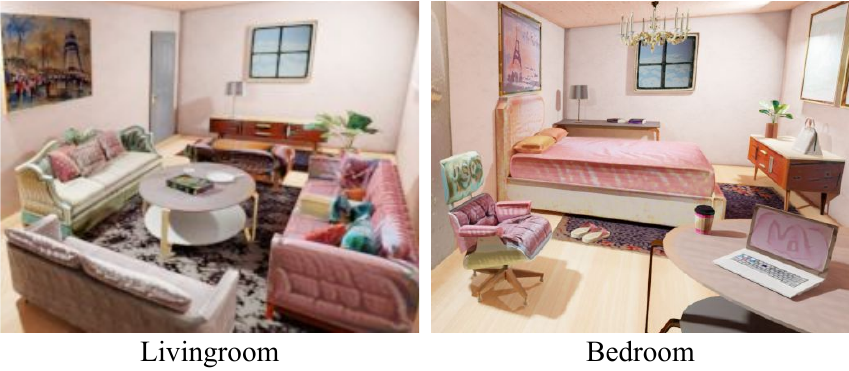}
\caption{Result for the different arrangement of objects.
We used target image number 3 in Fig.~\ref{fig:scene results} in both spaces.}
\vspace{-0.9em}
\label{fig:various room}
\end{figure}

\begin{table}[h!]

\centering
\resizebox{ \linewidth}{!}{
\begin{tabular}{lcccc}

\hline

& \multicolumn{1}{c}{\textit{-retrieval}} & 
\multicolumn{1}{c}{\textit{-hist,glo}} & 
\multicolumn{1}{c}{\textit{-detail}} & 
\multicolumn{1}{c}{Ours}\\

\hline

(Q1): Realistic & $2.23(\pm0.48)$ & $3.68(\pm0.49)$ & $\underline{4.02}(\pm0.49)$  &  $\textbf{4.22}(\pm0.42)$   \\


(Q2): Color & $2.09(\pm0.48)$ & $2.63(\pm0.59)$ & $\textbf{3.95}(\pm0.47)$ & $\underline{3.86}(\pm0.49)$  \\
\hline

\end{tabular}
}

\caption{Results of user study for scene stylization
}

\label{tab:scene_userstudy}
\vspace{-1.3em}
\end{table}

\begin{figure}[htbp!]
\centering
\includegraphics[width=\linewidth]{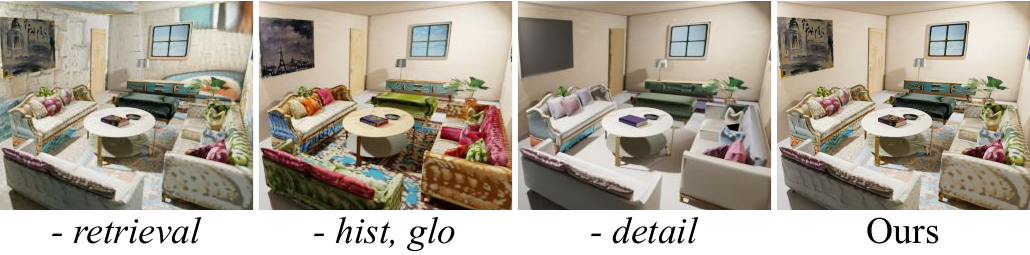}
\caption{Ablation results for the scene.
We used target image number 1 in Fig.~\ref{fig:scene results} in all spaces.}
\label{fig:scene_ablation}
\vspace{-0.5em}
\end{figure}

We also provide users' evaluation of the quality of our stylization in Table~\ref{tab:scene_userstudy}.
Users evaluate the results by answering the following questions: (Q1) `How realistic is the output result?’ (Q2) `How similar is the color distribution of the scene to the given images?’
Users assess the overall quality of the outputs, and how well they match the target image.
Since there are no previous works that use the same setting, we compare the results of ablated versions:
\textit{-retrieval} replaces the separate texture retrieval module (Sec.~\ref{sec:structure}) with a stylization network of objects for structure components;  
\textit{-hist,glo} removes the color loss and the global clip loss for the base color assignment (Sec.~\ref{sec:object_base}); and \textit{-detail} removes the detailed stylization step for objects (Sec.~\ref{sec:object_detail}).
The responses of (Q1) indicate that Text2Scene generates high-quality scenes and each of the components plays a crucial role to achieve reality.
The effect of texture retrieval is the most prominent.
The color distribution results (Q2) indicate that the base color assignment is critical.
The added local details are more important for the realism of the results.
Figure~\ref{fig:scene_ablation} shows exemplar images of the ablated versions used for the user study.

\paragraph{GPU Cost and Scalability}
Text2Scene first assigns base colors to discovered parts for all objects in the scene, and generates details of objects individually.
The cardinality for the base color assignment is only a few hundred and it does not require much memory and allows us to consider the whole scene while expanding the scale.
The most memory-intense process is the detail generation using a neural network, which only processes a single object at a time.
Therefore the entire process can be trained only on a single 11 GB GPU, making it an accessible tool for casual users.
In a single GPU, the base color allocation of the entire space takes 5 hours, and learning the details for each object takes 10 minutes.

\paragraph{Limitations} 
While our approach results in scene stylization, the pipeline separately handles individual objects after the base color assignment.
This is a practical choice for scalability but may lack an understanding of the context of the entire space.
Instead, we rely on the text description to weakly bind the objects into a similar style.
We can design the pipeline to receive an additional input with a texture map or a lightweight network and extend our model to better observe the holistic scene context within a limited GPU memory.
Also, our pipeline requires a class label or optional text description per object, which can be further automated.

\section{Conclusions}

We introduce Text2Scene, a novel framework to generate a texture for 3D scenes.
Our hierarchical framework can handle a variety of objects, including highly detailed textures for objects such as book or paintings.
By leveraging the representation power of pre-trained CLIP, the framework does not require any 3D datasets with texture or part annotation.
Given the 3D mesh models and the class labels or text descriptions of objects, our framework easily produces stylized results by picking a target image and a simple text description.
We hope that Text2Scene to facilitate the automatic interior recommendation or realistic virtual space generation.

\paragraph{Acknowledgements}
This work was supported by the National Research Foundation of Korea(NRF) grant funded by the Korea government(MSIT) (No. RS-2023-00208197) and Institute of Information \& communications Technology Planning \& Evaluation (IITP) grant funded by the Korea government(MSIT) (No.2021-0-02068, Artificial Intelligence Innovation Hub).
Inwoo Hwang is supported by Hyundai Motor Chung Mong-Koo Foundation. 
Young Min Kim is the corresponding author.

\clearpage
{\small
\bibliographystyle{ieee_fullname}
\bibliography{egbib}
}
\clearpage

\appendix

In the supplementary materials, we elaborate the implementation details for our Text2Scene (Sec.~\ref{sec:suppl_details}), the impact of seed (Sec.~\ref{sec:suppl_seed}), motivation of global context (Sec.~\ref{sec:global context}) additional visual results (Sec.~\ref{sec:suppl_additional_part}, Sec.~\ref{sec:compare_other}, Sec.~\ref{sec:suppl_additional_style}) and through algorithm (Sec.~\ref{sec:algorithm}).

\section{Implementation details}
\label{sec:suppl_details}
\subsection{Network Architecture}

The input to the LNSF is the positional encoding of the 3D coordinate, coefficients of the eigenfunctions of mesh, and part segment id, and the output is the color of the vertex.
Each vertex $p\in \mathbb{R} ^{3}$ is mapped to a 256-dimensional Fourier feature applying by $ \gamma ( p) =\left[ \cos \left( 2\pi B p\right) ,\sin \left( 2\pi B p\right) \right] $ where $B$ is randomly sampled from $\mathcal{N}\left( 0, 5 ^{2}\right)$.
Also, the coefficients of the eigenfunctions corresponding to the top 128 eigenvalues are used to reflect the intrinsic geometry. 
The neural network for LNSF consists of five 256-dimensional layers.
For activation, $\tanh$ is used for the last layer, and ReLU is used for others.
In addition, the weight of the final layer is set to zero; thus, the colors of all vertices meshes are initially set to a constant color.

For part-aware geometric deformation in Sec.~{\color{red}3.2.3}, we branch the last layer into two while having the first four layers in common, and each last layer outputs the color and displacement respectively~\cite{Michel2021text2mesh}.

\subsection{Training Details}

For all experiments, we use the Adam optimizer~\cite{Kingma:2015:Adam} with an initial learning rate of $5\times 10^{-4}$ and the learning rate decay factor as 0.9 for every 100 iterations.
We sample camera poses on a hemisphere with a radius $r$.
The viewing angles are sampled from Gaussian distribution with $\sigma=\pi/4$ around the front view of objects within the elevation angle of $\left[10^{\circ},80^{\circ}\right]$  and the azimuth angle of $\left[0^{\circ},360^{\circ}\right]$.
As mentioned in Sec.~{\color{red}3.2.3}. for detailed stylization, we apply random perspective transformations and random cropping and observe $10-20 \%$ of the original rendered image.
A color range of each point is set to 0 to 1 by adding half of the output of LNSF through $\tanh$ with the gray color $[0.5, 0.5, 0.5]$.

For parameters, we set $r=2.0$ for rendering, $\lambda _{1}=0.2$ for structure stylization and $\lambda_{th}= 3.0$ for merge segments.
Also, we set $\lambda _{2},\lambda _{3}$ as $0.2$ for base color assignment and $\alpha=0.3$ for detailed stylization.
For all experiments, we use a pre-trained CLIP learned with a ViT-B/32 backbone~\cite{2021CLIP}.

\subsection{Design Choice for Structure Stylization}

\begin{figure}[h!]
\centering
\includegraphics[width=0.95\linewidth]{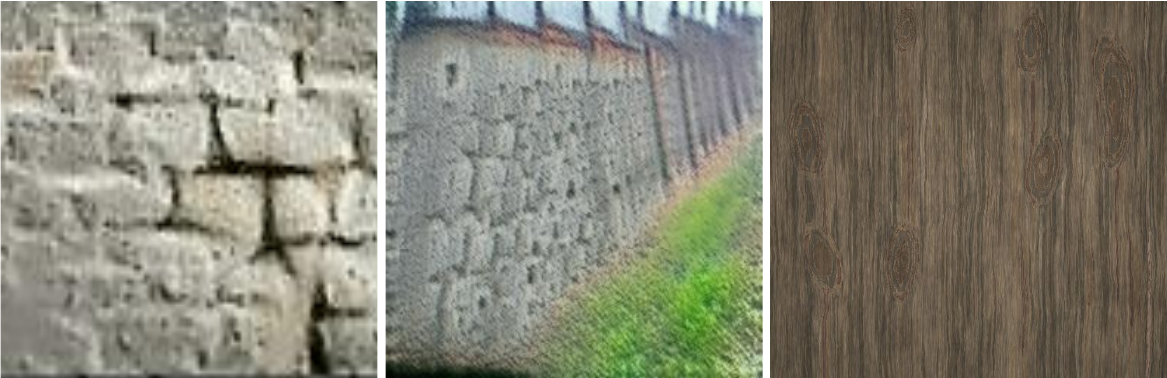}
\caption{Wall texture generation through MLP, CNN, and MATch~\cite{shi2020match}. 
Generation with CLIP embedding results in creating texture with multiple objects and perspective distortions, which is not appropriate for texture patterns for structural components with a flat geometry.
The text `a wall' is used for CLIP.}
\label{fig:wall artifact}
\end{figure}

In Sec.~{\color{red}3.1}, we stylize structure by retrieving texture from a pre-defined texture set using MATch~\cite{shi2020match}.
Here, we show undesirable artifacts when the structural components are not separately handled and were created using CLIP~\cite{2021CLIP}.
We generate walls directly from the designed MLP or optimize the weights of a CNN that translates a fixed random noise $z$ to an output image~\cite{deepimageprior,FeatureVisualization}.
As shown in Fig.~\ref{fig:wall artifact}, we observe various artifacts, such as a number of bricks or grass.
On the other hand, MATch~\cite{shi2020match} successfully samples a candidate of structure components.

\section{Impact of Seed, Limitation and Diversity}
\label{sec:suppl_seed}

\begin{figure}[h!]
\centering
\includegraphics[width=0.9\linewidth]{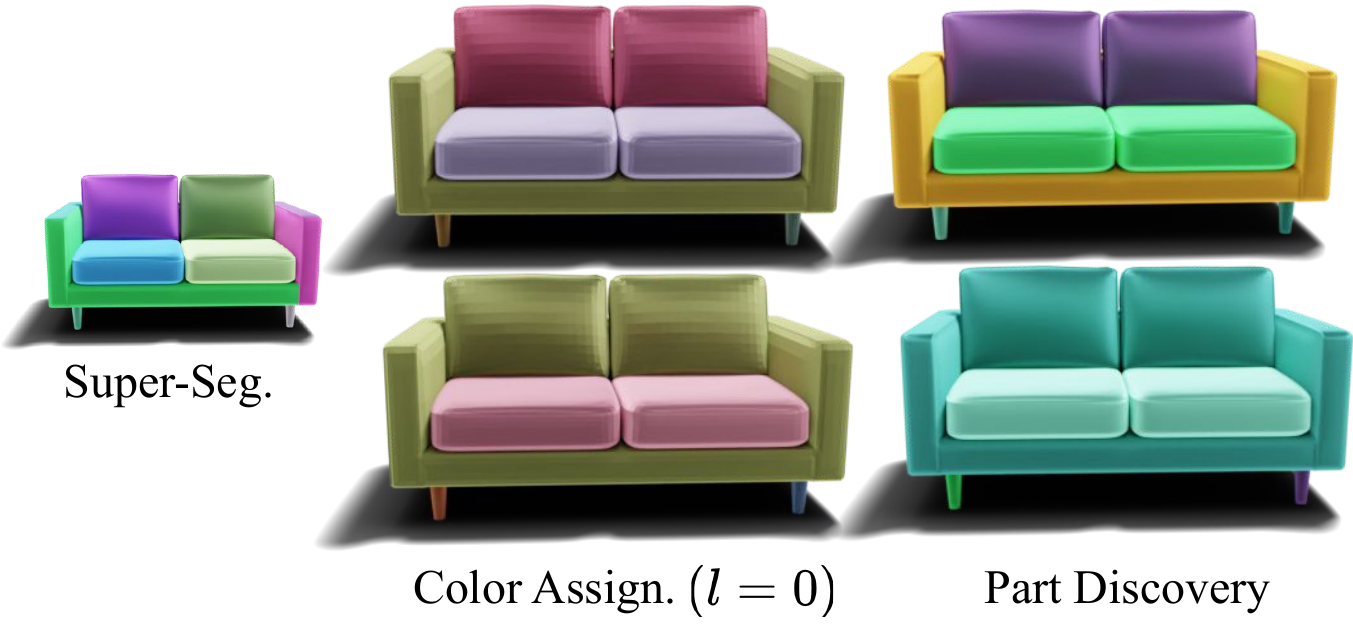}
\caption{From the same initial super segments, different parts can be observed through random seeds.}
\label{fig:part_failed}
\end{figure}

From the same text prompt, CLIP can generate different 2D images, and this property extends when we use CLIP embedding to discover part information or stylize 3D assets.
The additional degree of freedom stems from the random seed, as observed with the convergence pattern of generation using CLIP.
Figure~\ref{fig:part_failed} shows different part discovery results from the same initial super segments.
Different part segments are sometimes merged into one based on the convergence condition and might fail to discover the correct part information.

\begin{figure}[h!]
\centering
\includegraphics[width=0.95\linewidth]{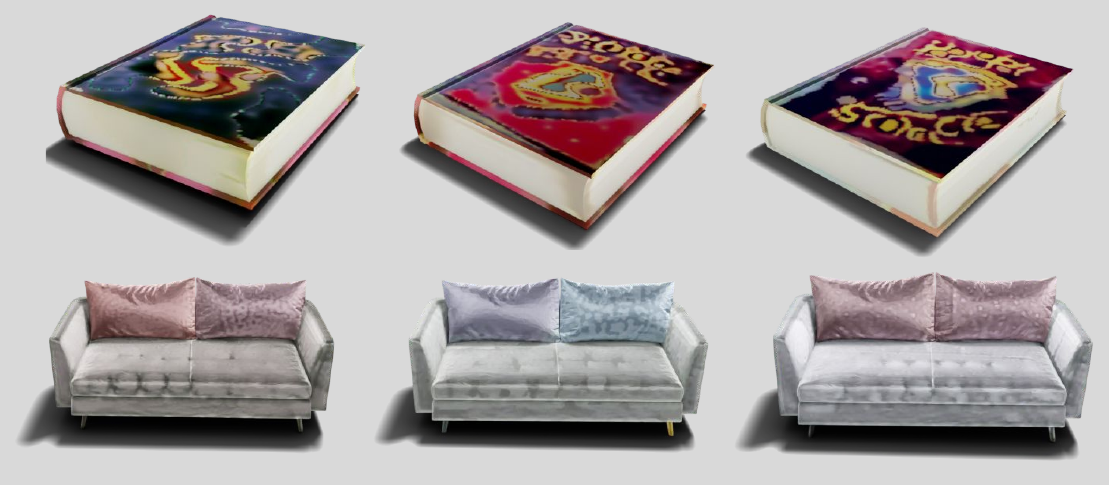}
\caption{From the same text prompt, we can generate diverse objects through different random seeds. `a book of Harry
Potter and the Sorcerer’s Stone' (top), and 'a sofa, minimal style' (bottom) is used for description, respectively.}
\label{fig:object seed}
\end{figure}

\begin{figure}[h!]
\centering
\includegraphics[width=0.95\linewidth]{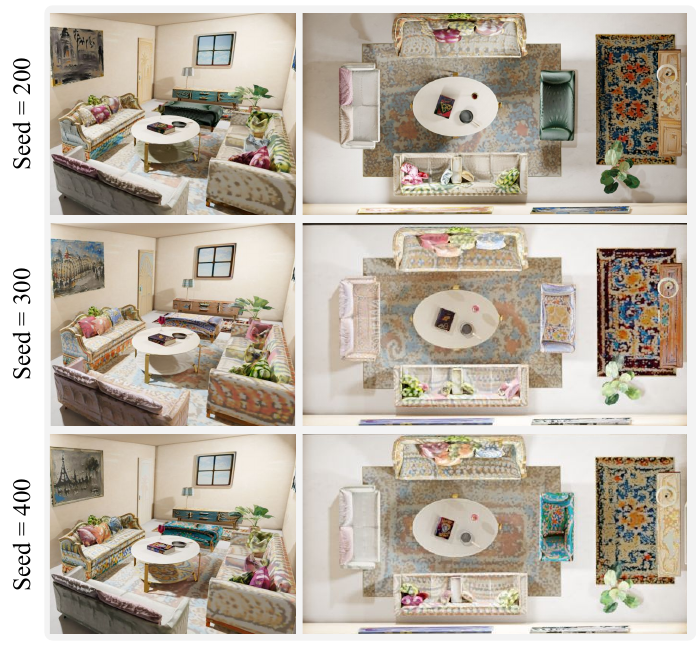}
\caption{Diverse scenes could be generated through different random seeds.}
\label{fig:scene seed}
\end{figure}

The different random seeds can result in diverse results after the part discovery.
Figure~\ref{fig:object seed} shows the stylization results of the same input text with different seeds.
Figure~\ref{fig:scene seed} shows stylized scenes of different seeds with the same input target image $I_t$ and the style description $t_s$.

\section{Motivation of Using Text for the Global Context}
\label{sec:global context}

We discuss our choice of the shared text description to enforce the clip loss in the structure stylization (Sec.~{\color{red}3.1}) and the part-level base color assignment for the object colors (Sec.~{\color{red}3.2.2}).

\begin{figure}[h!]
\centering
\includegraphics[width=0.95\linewidth]{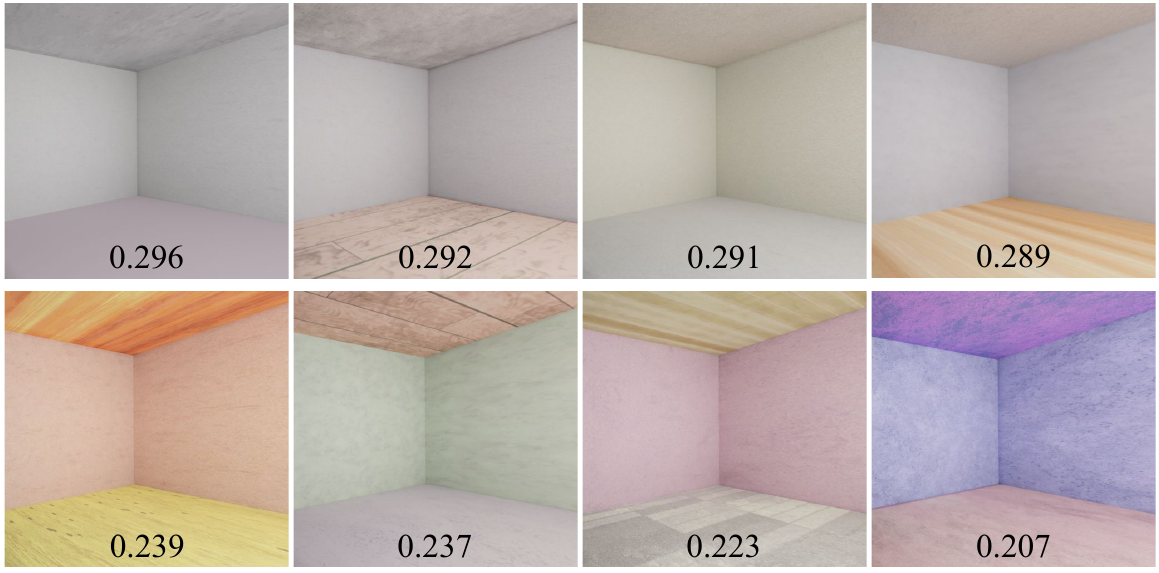}
\caption{$\mathcal{L}_\text{clip}\left( I_{s},T_{s}\right)$ with the renderings of structures from randomly sampled choices of textures.}
\label{fig:structure_text}
\end{figure}

For the structure stylization, we use an additional text prompt $T_s$, `a structure of a room' to provide the context with the clip loss.
Figure~\ref{fig:structure_text} shows randomly generated structure and their resulting clip scores against the text prompt $T_s$, $\mathcal{L}_\text{clip}\left( I_{s},T_{s}\right)$.
The samples indicate that the resulting clip scores, to some extent, reflect how natural the texture choices are.
Therefore, we can exclude unnatural stylization of the structural components with the simple text.

\begin{figure}[h!]
\centering
\includegraphics[width=0.95\linewidth]{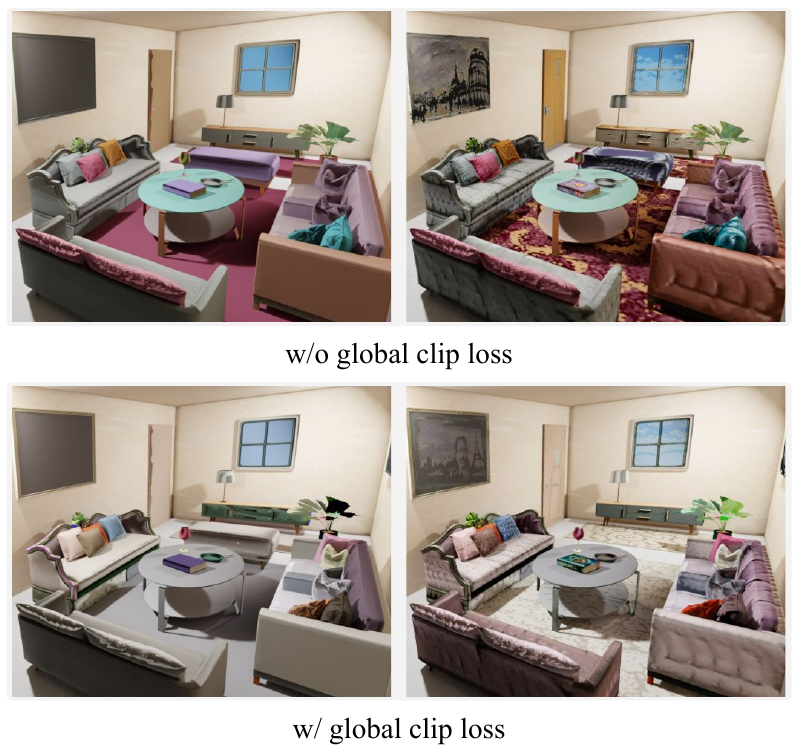}
\caption{Base color assignment results (left), and the final results with additional learned details (right). With the global clip loss, more harmonious base colors are assigned from which additional details are generated. The color histogram loss against the target image is not used.}
\label{fig:global_text}
\end{figure}

For objects within the scene, we assign part-level base colors with the text prompt $T$, which represents the types of the scene, such as `a bedroom', and an additional global signal $\mathcal{L}_\text{clip}\left( I,T\right)$.
Figure~\ref{fig:global_text} shows the results when each object is independently stylized using only the object clip loss compared to stylization with the additional global clip loss at the base color assignment step.
The color loss between the target image, $\mathcal{L}_\text{hist}\left( I,I_{t}\right)$  is not used.
With the global clip loss, we learn the assignments of the scene with harmonious colors of the scene-level rendering. 
Also, as shown in the results of the user study in the main paper (Table.~{\color{red}1}), we generate a more realistic texture for the holistic scene with the color histogram of the target image in addition to the global clip loss.

\section{Additional Results of Part Discovery}
\label{sec:suppl_additional_part}

\begin{figure}[h]
\centering
\includegraphics[width=\linewidth]{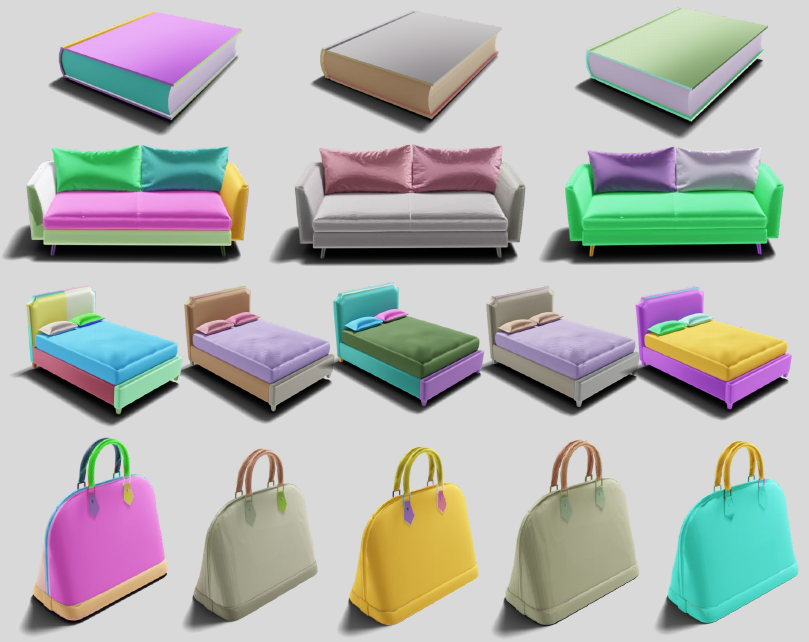}
\caption{Intermediate results of part discovery. From the initial super-segments (left), we show segment $\{s_{ik}^{l}\}$ and assigned color $c(s_{ik}^{l})$ at $l^\text{th}$ \text{iteration} and finally show the part discovery results (right) for each row. Empirically the process converges within two iterations. Random colors are assigned to each segment.}
\label{fig:part discovery visual}
\end{figure}

\begin{figure}[h]
\centering
\includegraphics[width=\linewidth]{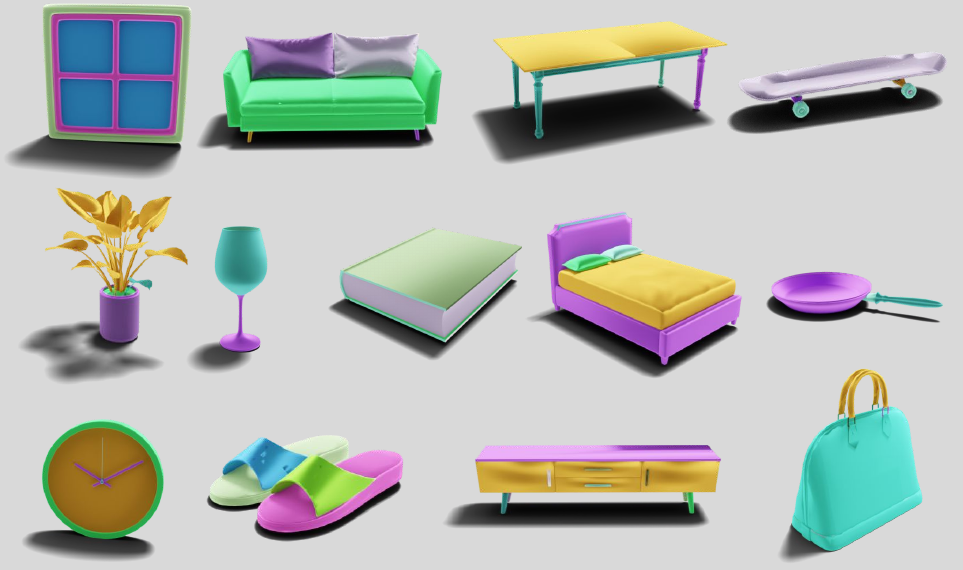}
\caption{Part discovery results for diverse categories of objects. Random colors are assigned to discovered parts.}
\label{fig:part example}
\end{figure}

Our realistic stylization greatly benefits from the stable part discovery to assign different textures. 
Figure~\ref{fig:part discovery visual} contains intermediate results of the entire pipeline, starting from the initial super-segments followed by iterations of merged segments until convergence. 
Figure~\ref{fig:part example} contains more results of the discovered parts of various objects.
It shows that the results correctly capture different parts that conventionally are colored with different materials or textures.

We observed a 76.6\% success rate for the part recovery without the help of any part dataset.
Note that there is no public dataset available for `texture parts', which are different from functional parts or semantic segmentation, and we evaluated with a manually annotated dataset.

\section{Comparison to Other Text-to-3D Methods}
\label{sec:compare_other}

\begin{figure}[h]
\centering
\includegraphics[width=\linewidth]{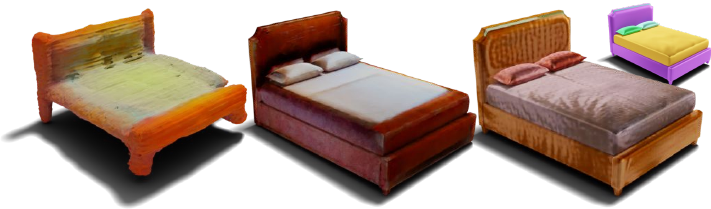}
\caption{Result from \cite{poole2022dreamfusion},~\cite{metzer2022latent}, and ours with discovered part in order.}
\label{fig:other3d}
\end{figure}

Parallel to our method, several methods create 3D contents from text input.
such as DreamFusion\cite{poole2022dreamfusion} and Latent-NeRF~\cite{metzer2022latent}.
DreamFusion\cite{poole2022dreamfusion} creates 3D contents in NeRF representation.
As shown in Figure~\ref{fig:other3d}, the created volumetric representation tends to be blurry for highly structured objects, such as beds, while we enjoy more explicit texture boundaries.
We also include a comparison against Latent-NeRF~\cite{metzer2022latent}, which supports the mode of updating the UV-texture map of a mesh using the score distillation loss proposed by \cite{poole2022dreamfusion}.
The color distribution generated with our method exhibits superior visual quality over their texture map as our results show clear boundaries obtained from the part discovery step.

\section{Additional Stylization Results}
\label{sec:suppl_additional_style}

\begin{figure*}[t]
\centering
\includegraphics[width=\linewidth]{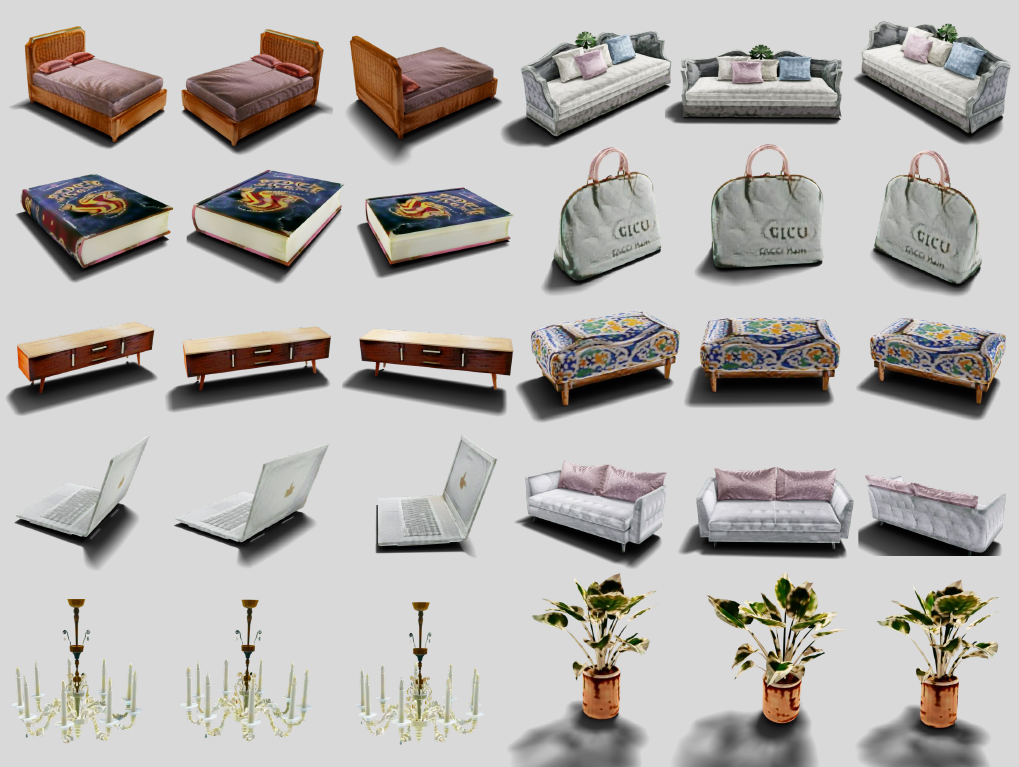}
\caption{We stylize whole 3D objects and provide a diverse view of stylized objects.}
\label{fig:diverse view}
\end{figure*}

\begin{figure*}[h]
\centering
\includegraphics[width=\linewidth]{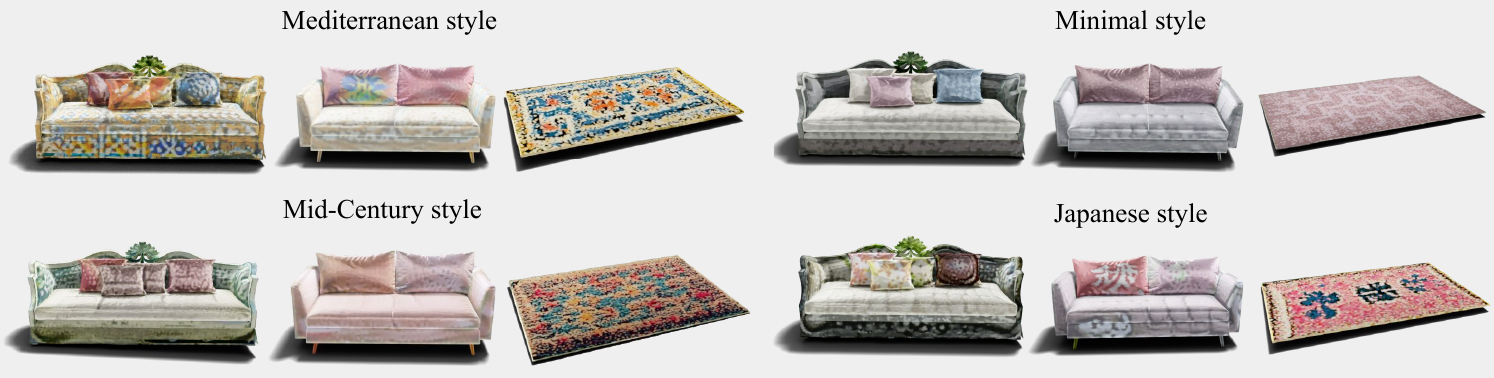}
\caption{Object stylization results in the specific text description.}
\label{fig:object style}
\end{figure*}

Although it was not obvious from the main manuscript, our stylized objects contain plausible texture even when observed from diverse points, as shown in Fig.~\ref{fig:diverse view}.
The individual objects can be weakly bound by the text description as shown in Fig.~\ref{fig:object style}. 

\begin{figure*}[ht]
\centering
\includegraphics[width=0.95\linewidth]{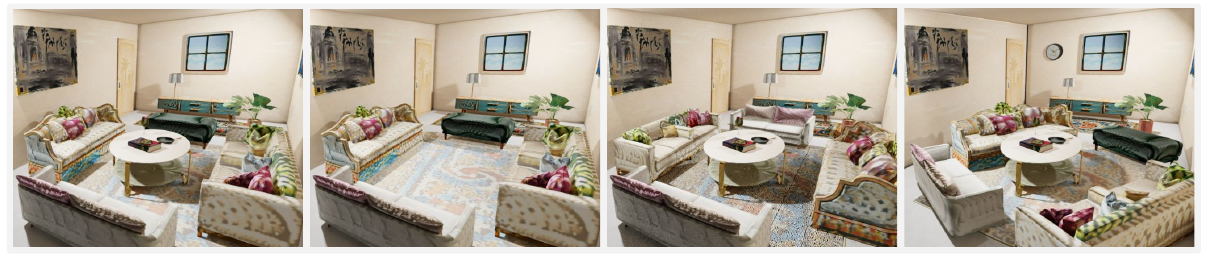}
\caption{From original scene (left), we can manipulate scene by object removal, replication and relocation.}
\label{fig:scene edit}
\end{figure*}

\begin{figure*}[ht]
\centering
\includegraphics[width=0.95\linewidth]{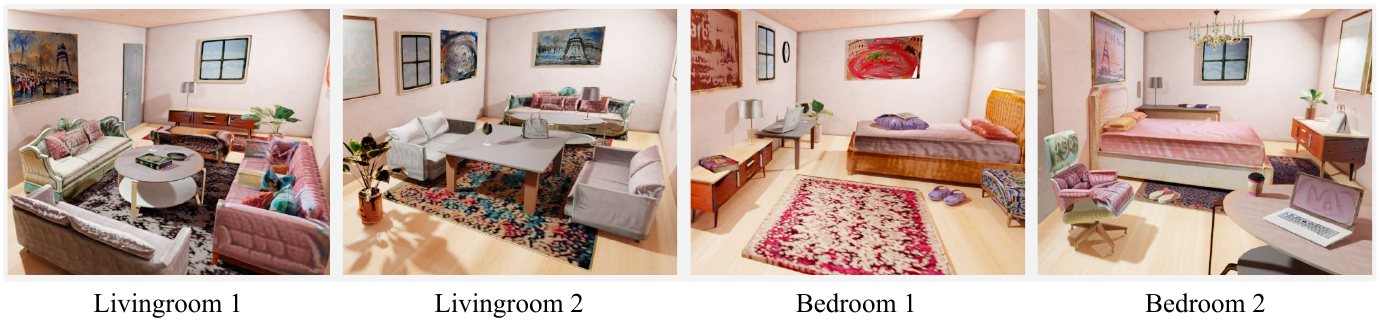}
\caption{Additional stylize results for various scenes. We validate our algorithms for four scenes.}
\label{fig:diverse scene}
\end{figure*}

\begin{figure*}[ht]
\centering
\includegraphics[width=\linewidth]{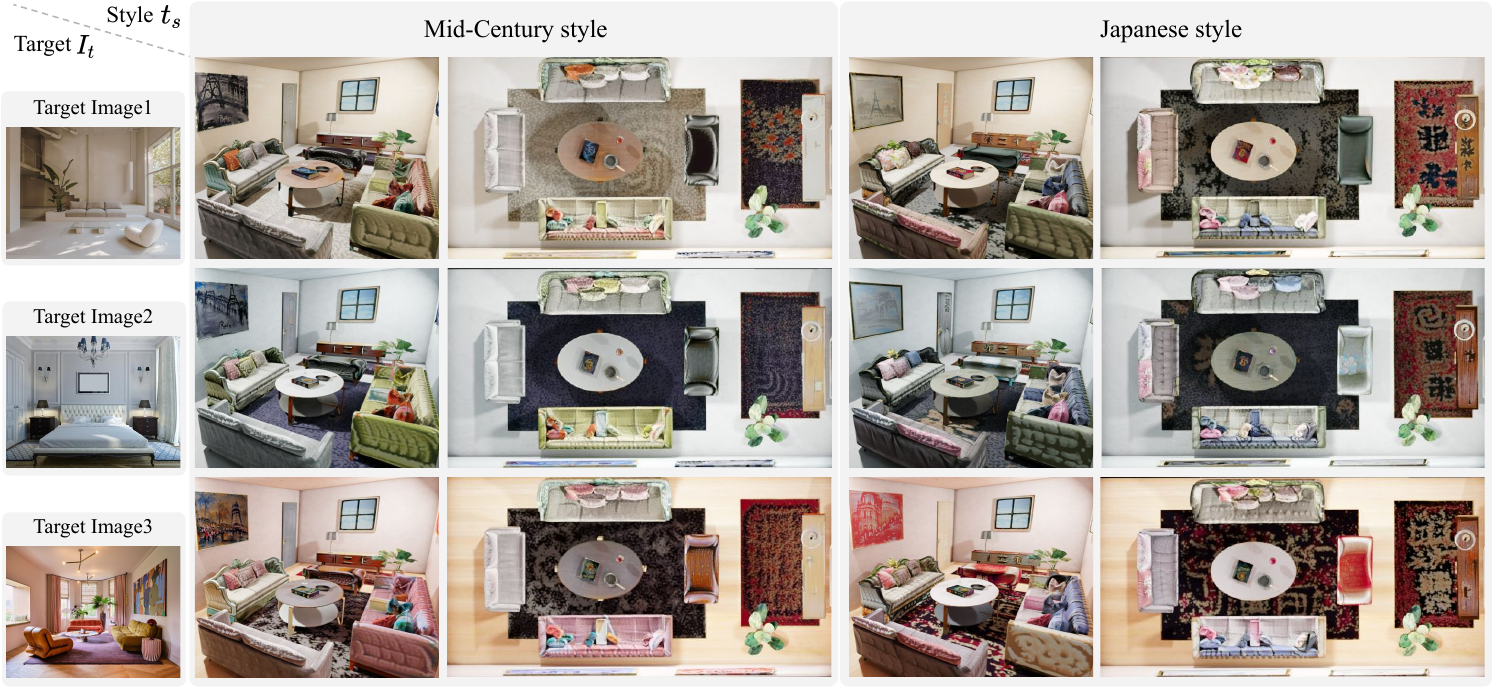}
\caption{Results of the same scene with different stylization, using different target images and style texts.}
\label{fig:scene style}
\end{figure*}

\begin{figure*}[h]
\centering
\includegraphics[width=\linewidth]{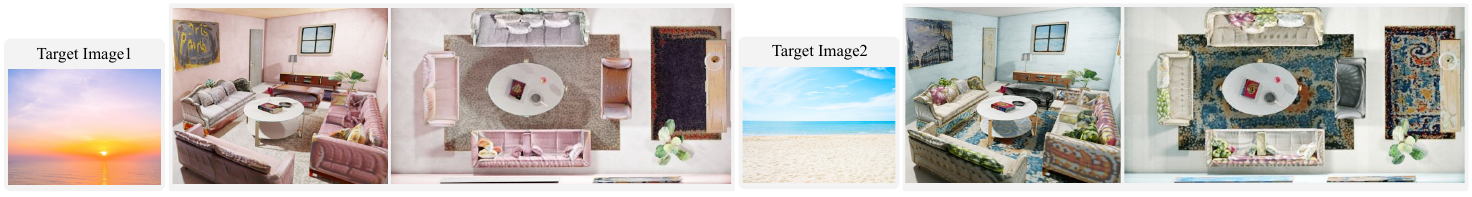}
\caption{Results of setting the target image as a natural photograph.}
\label{fig:scene nature}
\end{figure*}

Since we stylize the entire object, we can easily manipulate the 3D scene through object removal, replication, or relocation (Fig.~\ref{fig:scene edit}).
Also, we show additional stylized results over various scenes (Fig.~\ref{fig:diverse scene}), or target image and style descriptions (Fig.~\ref{fig:scene style}).
Finally, for the scene stylization, our target image can be replaced to natural photographs (Fig.~\ref{fig:scene nature}).

\section{Text2Scene: Algorithm}
\label{sec:algorithm}
We provide the algorithm to describe the flow of the entire pipeline in Algorithm~\ref{alg:through_algorithm}.

\clearpage
\begin{algorithm*}[t]
    \caption{Overall pipeline of Text2Scene}

    \begin{algorithmic}[1]
    \Require {\small 3D scene $\mathcal{S}=\{\mathcal{W}, \mathcal{O}\}$ where $\mathcal{W}$ is a structure components and $\mathcal{O}=\{\mathbf{M}_{i}\}$ is a set of 3D mesh objects,
    
    Corresponding class labels and optionally have text descriptions for each $\mathbf{M}_{i}$,
    
    Target image $\mathbf{I}_t$, specific appearance style description $\mathbf{t}_{s}$}
    \Ensure {\small Stylized 3D Scene $\mathcal{S}$ reflecting object specific information and given conditions}
    \vspace{2mm}

    \noindent \grc{\small \# {Structure Stylization $\mathcal{W}$} (Sec. 3.1.)}\vspace{1mm}
    \State $Score = \emptyset{}$
    \For{$\text{i}=1,2,\dots,N$}
        \State $\mathcal{W} \gets \verb|set_texture|$
        \Comment{\small Sample texture from texture set}
        \State $\mathbf{I}_{s} \gets \verb|render|(\mathcal{W})$
        \State $Score = Score \cup \{criteria(\mathbf{I}_{s}, \mathbf{I}_{t}, \mathbf{T}_{s})\}$
    \EndFor
    \State $[\mathcal{W}^{*}, \underline{\hspace{3mm}}\,] \gets \verb|top-1|[Score]$
    \vspace{2mm}

    \noindent \grc{\small \# {Part Discovery for each 3D object (Sec. 3.2.1.)}} 
    \vspace{1mm}
    
    \For{$\text{i}=1,2,\dots,|\mathcal{O}|$}
        \State $\{\mathbf{s}_{ik}^{0}\}  \gets \verb|superseg|(\mathbf{M}_{i})$
        \Comment{\small Initial super-segments}
        \State $l=0$
        
        \While {num of segments does not decrease}
            \State Set $\mathbf{c}(\mathbf{s}_{ik}^{l})$ as grey
            \State Generate a graph $\mathcal{G}_{i}^{l}$
            \For{$\text{iter}=1,2,\dots,L$}
                \State $\mathbf{I}_{M_i} \gets \verb|render|(\mathbf{M}_{i})$
                \State $\mathcal{L} \gets \mathcal{L}_\text{clip}(\mathbf{I}_{M_i}, \mathbf{T}_{i,c})$
                \State Update $\mathbf{c}(\mathbf{s}_{ik}^{l})$
            \EndFor
            \State Update graph $\mathcal{G}_{i}^{l}$
            \State $\{\mathbf{s}_{ik}^{l+1}\} \gets \verb|merge|(\{\mathbf{s}_{ik}^{l}\})$
            \Comment{\small Merge segments}
            \State $l=l+1$
        \EndWhile
    \EndFor
    \State Discovered Part $\{\mathbf{s}_{ik}\}$
    \vspace{2mm}
    
    \noindent \grc{\small \# {Part-level Base Color Assignment  (Sec. 3.2.2.) }}\vspace{1mm}
    \State $\forall \mathbf{M}_{i} \in \mathcal{O}, \,\, $ set $ \mathbf{c}(\mathbf{s}_{ik})$ as grey
    \For{$\text{iter}=1,2,\dots,L$}
        \State $\mathbf{I} \gets \verb|render|(\mathcal{W}^{*}, \mathcal{O})$
        \State $\mathbf{I}_{M_i} \gets \verb|render|(\mathbf{M}_{i}), \,\,\, \forall  \mathbf{M}_{i} \in \mathcal{O}$
        \State $\mathcal{L} \gets \mathcal{L}_\text{color,scene} + \mathcal{L}_\text{clip,scene}$
        \State Update $\mathbf{c}(\mathbf{s}_{ik}), \,\,\, \forall  \mathbf{M}_{i} \in \mathcal{O}$
    \EndFor
    \State Part-level Assigned Color $\mathbf{c}(\mathbf{s}_{ik})$
    
    \vspace{2mm}
    
    \noindent \grc{\small \# {Detailed Stylization  (Sec. 3.2.3.)}}\vspace{1mm}
    
    \For{$\text{i}=1,2,\dots,|\mathcal{O}|$}
        \For{$\text{iter}=1,2,\dots,L$}
            \State $\mathbf{M}_{i} \gets \verb|update_local_texture|( \mathcal{F}_{i,\theta}(\mathbf{M}_{i}))$
            \Comment{\small Local Neural Style Field (LNSF)}
            \State $\mathbf{I}_{M_i} \gets \verb|render|(\mathbf{M}_{i})$
            \State $\mathcal{L} \gets \mathcal{L}_\text{clip}( I_{M_{i}},T^{+}_{i})$
            \State Update LNSF $\mathcal{F}_{i}$ parameters
        \EndFor
    \EndFor
    \State Stylized Object  $\mathbf{M}_{i}$
    
    \end{algorithmic}
    \label{alg:through_algorithm}
    
\end{algorithm*}   

\clearpage

\end{document}